\documentclass{article}
\usepackage{spconf,amsmath,graphicx}
\usepackage{hyperref}

\title{Prediction of Deep Ice Layer Thickness Using Adaptive Recurrent Graph Neural Networks}
\name{Benjamin Zalatan$^1$, Maryam Rahnemoonfar$^{1,2,*}$ \thanks{$^*$Corresponding author (maryam@lehigh.edu).}}

\address{
$^1$Department of Computer Science and Engineering, Lehigh University, PA, USA\\
$^2$Department of Civil and Environmental Engineering, Lehigh University, PA, USA
}

\begin{document}
%\ninept
%
\maketitle
\begin{abstract}
%As we deal with the effects of climate change and the increase of global atmospheric temperatures, the accurate tracking and prediction of ice layers within polar ice sheets grows in importance. Airborne radar sensors, such as the Snow Radar, are able to capture these deep ice layers, and by proxy, historic accumulation rate patterns at a large scale. This allows researchers to study the effects of climate change on polar regions via their precipitation and run-off. The Snow Radar's use of an ultra-wide bandwidth enables a fine vertical resolution that helps in capturing internal ice layers. Given the amount of snow accumulation in recent years gathered through this radar data, we propose a machine learning model that uses adaptive, recurrent graph convolutional networks to predict historic snow accumulation by way of the thickness of deep ice layers. We found that our model performs better than previous models and equivalent nonadaptive and nontemporal models.

As we deal with the effects of climate change and the increase of global atmospheric temperatures, the accurate tracking and prediction of ice layers within polar ice sheets grows in importance. Studying these ice layers reveals climate trends, how snowfall has changed over time, and the trajectory of future climate and precipitation. In this paper, we propose a machine learning model that uses adaptive, recurrent graph convolutional networks to, when given the amount of snow accumulation in recent years gathered through airborne radar data, predict historic snow accumulation by way of the thickness of deep ice layers. We found that our model performs better and with greater consistency than our previous model as well as equivalent non-temporal, non-geometric, and non-adaptive models.
\end{abstract}
\section{Introduction}
\label{sec:intro}

As global atmospheric temperatures rise and climate trends shift, there has been a growing importance placed upon accurately tracking and predicting polar snowfall over time. A precise understanding of the spatiotemporal variability in polar snow accumulation is important for reducing the uncertainties in climate model predictions, such as prospective sea level rise. These snowfall trends are revealed through the internal ice layers of polar ice sheets, which often represent annual isochrones and relay information about the climate at that location during the corresponding year, similar to rings on a tree. The tracking and forecasting of these internal ice layers is also important for calculating snow mass balance, extrapolating ice age, and inferring otherwise difficult-to-observe processes.

Measurements of ice layer mass balance are traditionally collected by drilling ice cores and shallow pits. However, capturing catchment-wide accumulation rates using these methods is exceedingly difficult due to their inherent sparsity, access difficulty, high cost, and depth limitations. Attempts to interpolate these in-situ measurements introduce further uncertainties to climate models, especially considering the high variability in local accumulation rate.

Airborne measurements using nadir-looking radar sensors has quickly become a popular complementary method of mapping ice sheet topography and monitoring accumulation rates with a broad spatial coverage and ability to penetrate deep ice layers. The Center for Remote Sensing of Ice Sheets (CReSIS), as part of NASA's Operation Ice Bridge, operates the Snow Radar \cite{snow-radar}, an airborne radar sensor that takes high-resolution echograms of polar ice sheets.

Recent studies involving graph convolutional networks (GCNs) \cite{gcn} have shown promise in spatiotemporal tasks such as traffic forecasting \cite{relevant_traffic_rgcn, traffic1, traffic2}, wind speed forecasting \cite{wgatlstm}, and power outage prediction \cite{poweroutagegnn}. In this paper, we propose a geometric deep learning model that uses a supervised, multi-target, adaptive long short-term memory graph convolutional network (AGCN-LSTM) \cite{gcnlstm, evolvegcn} to predict the thicknesses of multiple deep ice layers at specific coordinates in an ice sheet given the thicknesses of few shallow ice layers. 

In our experiments, we use a sample of Snow Radar flights over Greenland in the year 2012. We convert this internal ice layer data into sequences of temporal graphs to be used as input to our model. More specifically, we convert the five shallow ice layers beneath the surface into five spatiotemporal graphs. Our model then performs multi-target regression to predict the thicknesses of the fifteen deep ice layers beneath them. Our model was shown to perform significantly better than previous models in predicting ice layer thickness, as well as better than equivalent non-geometric, non-adaptive, and non-temporal models.

\section{Related Work}
\label{sec:relatedwork}

\subsection{Automated Ice Layer Segmentation}

In recent years, automated techniques have been developed to track the surface and bottom layers of an ice sheet using radar depth sounder sensors.
Tracking the internal layers, however, is more difficult due to the low proximity between each layer, as well as the high amount of noise present in the echogram images. Due to its exceptional performance in automatic feature extraction and image segmentation tasks, deep learning has been applied extensively on ice sheet echograms in order to track their internal layers \cite{rahnemoonfar2021deep, varshney2020deep, yari2019smart, yari2020multi}. \cite{yari2019smart} used a multi-scale contour-detection convolutional neural network (CNN) to segment the different internal ice layers within Snow Radar echogram images. In \cite{rahnemoonfar2021deep}, the authors trained a multi-scale neural network on synthetic Snow Radar images for more robust training. A multi-scale network was also used in \cite{yari2020multi}, where the authors trained a model on echograms taken in the year 2012 and then fine tuned it by training on a small number of echograms taken in other years. \cite{varshney2020deep} found that using pyramid pooling modules, a type of multi-scale architecture, helps in learning the spatio-contextual distribution of pixels for a certain ice layer. The authors also found that denoising the input images improved both the model's accuracy and F-score. While these models have attempted to segment Snow Radar echogram images, none have yet attempted to predict deep ice layer thicknesses with only information about shallow ice layers.

\subsection{Graph Convolutional Networks}

Graph convolutional networks have had a number of applications in a vast array of different fields. In the field of computer vision, recurrent GCNs have been used to generate and refine ``scene graphs'', in which each node corresponds to the bounding box of an object in an image and the edges between nodes are weighted by a learned ``relatedness'' factor \cite{scenegraph1, scenegraph2}. GCNs have also been used to segment and classify point clouds generated from LiDAR scans \cite{lidar1, lidar2}. Recurrent GCNs have been used in traffic forecasting, such as in \cite{relevant_traffic_rgcn}, where graph nodes represented traffic sensors, edges were weighted by the physical distance between sensors, and node features consisted of the average detected traffic speed over some period of time.

Some existing graph-based weather prediction models, such as \cite{wgatlstm} and \cite{wgcnlstm}, have tested models in which edge weights are defined as learnable parameters rather than static values. This strategy allowed the models to learn relationships between nodes more complex than simple geographic distance, and was shown to improve performance at the expense of increased computational complexity.

In our previous study published at the 2023 IEEE Radar Conference \cite{radarpaper}, we used a GCN-LSTM to predict the thicknesses of shallow ice layers using the thicknesses of deep ice layers. Our results were reasonable, usually lying within 5 pixels of the ground-truth, and we found that GCN-LSTM performed better and with more consistency than equivalent non-temporal and non-geometric models. While this previous model had a similar objective to the model described in this paper, it was far less complex, did not include learned adjacency, and attempted to predict the thicknesses of shallow ice layers rather than deep ice layers.

\section{Dataset}
\label{sec:dataset}

In this study, we use the Snow Radar dataset made public by CReSIS as part of NASA's Operation Ice Bridge. The Snow Radar operates from 2-8 GHz and is able to track deep ice layers with a high resolution over wide areas of an ice sheet. The sensor produces a two-dimensional grayscale profile of historic snow accumulation over consecutive years, where the horizontal axis represents the along-track direction, and the vertical axis represents layer depth. Pixel brightness is directly proportional to the strength of the returning signal. Each of these grayscale echogram profiles has a width of 256 pixels and a height ranging between 1200 and 1700 pixels. Each pixel in a column corresponds to approximately 4cm of ice, and each echogram image has an along-track footprint of 14.5m. Accompanying each image are vectors that provide positional data (including geographic latitude and longitude) of the sensor for each column. In order to gather ground-truth thickness data, the images were manually labelled in a binary format where white pixels represented the tops of each firn layer, and all other pixels were black. Thickness data was extracted by finding the distance (in pixels) between each white pixel in a vertical column.

We focus on radar data captured over Greenland during the year 2012. Since each ice layer often represents an annual isochrone, we may refer to specific layers by their corresponding year (in this case, the surface layer corresponds with the year 2012, the layer below it 2011, and so on). In order to capture a sufficient amount of data, only echogram images containing a miniumum of 20 ice layers were used (five feature layers and fifteen predicted layers). Five and fifteen feature and predicted layers, respectively, were chosen in order to maximize the number of usable images while maintaining a sufficient number of experimental layers. This restriction reduced the total number of usable images down to 703. Five different training and testing sets were generated by taking five random permutations of all usable images and splitting them at a ratio of 4:1. Each training set contained 562 images, and each testing set contained 141 images.

\section{Methods}
\label{sec:methods}

\begin{figure*}
    \centerline
    {
        \includegraphics[width=\textwidth]{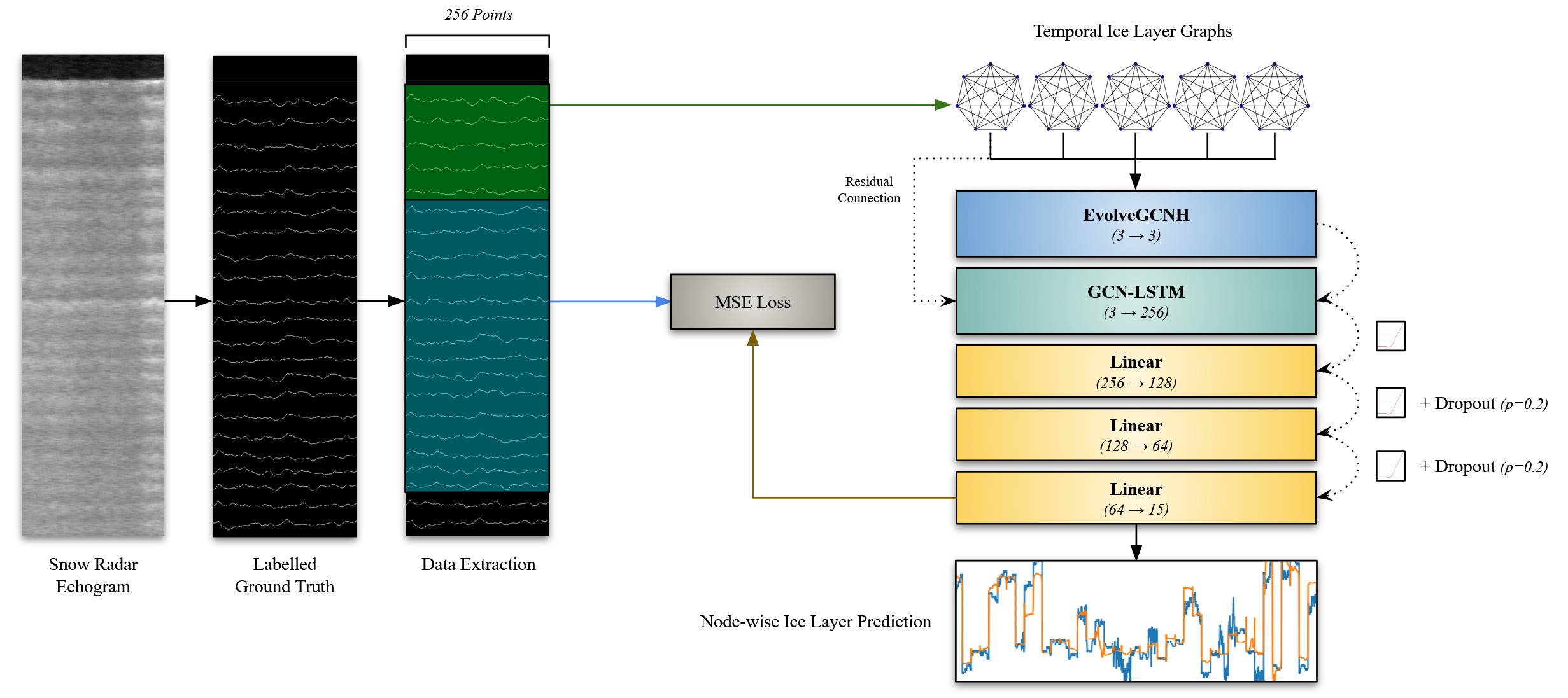}
    }
    \caption{Architecture of the proposed model.}
    \label{fig:arch}
\end{figure*}

\subsection{Graph Convolutional Networks}

Traditional convolutional neural networks use a matrix of learnable weights, often referred to as a kernel or filter, as a sliding window across pixels in an input image. The result is a higher-dimensional representation of the image that automatically extracts image features that would otherwise need to be identified and inputted manually. Graph convolutional networks apply similar logic to graphs, but rather than using a sliding window of learned weights across a matrix of pixels, GCN performs weighted-average convolution on each node's neighborhood to automatically extract features that reflect the structure of a graph. The size of the neighborhood on which convolution takes place is dictated by the number of sequential GCN layers present in the model (i.e. $K$ GCN layers results in $K$-hop convolution). In a sense, GCNs are a generalized form of CNNs that enable variable degree.

A special form of GCN, known as adaptive GCN (or AGCN), define edge weights within an input graph as learnable parameters rather than predefined constants. In certain cases, this may increase model performance if relationships between nodes are more advanced than those specified by the input. In the case of our model, we route the graphs through an EvolveGCNH layer \cite{evolvegcn} prior to entering the GCN-LSTM layer.

EvolveGCNH is a version of EvolveGCN that behaves similarly to a traditional GCN, but treats its learned weight matrix as a temporal hidden state that, through use of a gated recurrent unit (GRU), implicitly adjusts the structure of input graphs by modifying node embeddings. The adjustment of the weight matrix at each forward pass is influenced by the previous hidden weight state as well as the node embeddings of the current input graph.

\subsection{Recurrent Neural Networks}

Recurrent neural networks (RNNs) are able to process a sequence of data points as input, rather than a single static data point, and learn the long-term relationships between them. Many traditional RNN structures have had issues with vanishing and exploding gradients on long input sequences. Long short-term memory (LSTM) \cite{lstmpaper} attempts to mitigate those issues by implementing gated memory cells that guarantee constant error flow. Applying LSTM to GCN using GCN-LSTM allows for a model to learn not only the relationships between nodes in a graph, but also how those relationships change (or persist) over time.

\subsection{Model Architecture}

Our model (see Figure \ref{fig:arch}) uses an EvolveGCNH layer to introduce adaptivity to input adjacency matrices. The resulting node matrix is used as the feature matrix for a GCN-LSTM layer with 256 output channels. This leads into three fully-connected layers: the first with 128 output channels, the second with 64 output channels, and the third with 15 output channels, each corresponding to one of the 15 predicted ice layer thicknesses. Between each layer is the Hardswish activation function \cite{hardswishpaper}, an optimized approximation of the Swish function that has been shown to perform better than ReLU and its derivatives in deep networks \cite{swishbetter}. Between the fully-connected layers is Dropout \cite{dropoutpaper} with p=0.2. We use the Adam optimizer \cite{adampaper} over 300 epochs with mean-squared error loss. We use a dynamic learning rate that halves every 75 epochs beginning at 0.01.

\begin{table*}
    \centering
    \caption{Results from the non-temporal, non-geometric, non-adaptive, and proposed models on the fifteen predicted annual ice layer thicknesses from 1992 to 2006. Results are shown as the mean $\pm$ standard deviation of the RMSE over five trials (in pixels).}
    \begin{tabular}{ | c | c | c | c | c |  } 
        \hline
        & LSTM & GCN & GCN-LSTM & AGCN-LSTM  \\ 
        \hline
        Total RMSE & $5.817 \pm 1.349$ & $3.496 \pm 0.509$ & $2.766 \pm 0.312$ & $\mathbf{2.712 \pm 0.179}$ \\ 
        \hline
    \end{tabular}
    \label{table:OverallResults}
\end{table*}

\subsection{Graph Generation}

Each ground-truth echogram image is converted into five graphs, each consisting of $256$ nodes. Each graph corresponds to a single ice layer for each year from 2007 to 2011. Each node represents a vertical column of pixels in the ground-truth echogram image and has three features: two for the latitude and longitude at that point, and one for the thickness of the corresponding year's ice layer at that point.

All graphs are fully connected and undirected. All edges are inversely weighted by the geographic distance between node locations using the haversine formula. For a weighted adjacency matrix $A$:

{\small
    \begin{equation*}
        A_{i, j} = \frac{1}{2\arcsin\bigg(\text{hav}(\phi_j - \phi_i) + \cos(\phi_i)\cos(\phi_j)\text{ hav}(\lambda_j - \lambda_i)\bigg)}
    \end{equation*}
}
where
{\small
\begin{equation*}
    \text{hav}(\theta) = \sin^2 \bigg(\frac{\theta}{2} \bigg)
\end{equation*}
}
$A_{i, j}$ represents the weight of the edge between nodes $i$ and $j$. $\phi$ and $\lambda$ represent the latitude and longitude features of a node, respectively. Node features of all graphs are collectively normalized using z-score normalization. Weights in the adjacency matrices of all graphs are collectively normalized using min-max normalization with a slight offset to prevent zero- and one-weight edges. Self-loops are added with a weight of two. While we use an EvolveGCNH layer to introduce learned adjacency, this predefined spatial adjacency matrix serves as the initial state of the learned adjacency matrix, and is also passed residually to the GCN-LSTM layer.

\section{Results}
\label{sec:results}

%\begin{table}
%    \caption{Key parameters of the Snow Radar used during data collection.} 
%    \begin{centering}
%        \label{table:SnowRadarParam}
%        \begin{tabular}{ll|l}
%        $\quad$ &$\quad$& Mean RMSE \\
%        \hline
%        LSTM &$\quad$& ... \\
%        GCN & $\quad$ & $2.949 \pm 0.129$ \\
%        GCN-LSTM &$\quad$& $2.758 \pm 0.303$ \\
%        AGCN-LSTM  &$\quad$& $\mathbf{2.650 \pm 0.178}$ \\
%        \end{tabular}
%    \end{centering}
%\end{table}

In order to verify that the temporal and adaptive aspects of the model serve to its benefit, we compared its performance with equivalent non-geometric, non-temporal, and non-adaptive models.

For the non-geometric model, the EvolveGCNH and GCN-LSTM layers are replaced by a single LSTM layer, and all node feature data is concatenated into a single, stacked feature vector. Since this model is non-geometric, no adjacency data is supplied. All other hyperparameters remain the same.

For the non-temporal model, the EvolveGCNH and GCN-LSTM layers are replaced by a single GCN layer. Rather than generating five independent graphs for each of the five shallow ``feature'' ice layers, we generate a single graph and concatenate the thickness features from all five graphs together. The rest of the model, including the adjacency matrix generation, is identical to the proposed model.

For the non-adaptive model, all hyperparameters remain the same, but the adaptive EvolveGCNH layer is removed. The rest of the model remains identical to the proposed model.

Over each trial, the root mean squared error (RMSE) was taken between the predicted and ground truth thickness values for each of the fifteen ice layers from 1992 to 2006 over all images in its corresponding testing set. The mean and standard deviation RMSE over all five trials are displayed in Table \ref{table:OverallResults}. The proposed AGCN-LSTM model consistently performed better than the baseline models in terms of mean RMSE.

\section{Conclusion}
\label{sec:conclusion}

In this work, we proposed a temporal, geometric, adaptive multi-target machine learning model that predicts the thicknesses of deep ice layers within the Greenland ice sheet (corresponding to the annual snow accumulation from 1992 to 2006, respectively), given the thicknesses of shallow ice layers (corresponding to the annual snow accumulation from 2007 to 2011, respectively). Our proposed model was shown to perform better and with more consistency than equivalent non-geometric and non-temporal models.

\subsection{Improvements and Generalizations}

While our model succeeds at predicting deep layer thicknesses with reasonable accuracy, there are opportunities for improvement and further generalization. For example, it may be possible to use radar data from multiple different years in order to adjust the model to predict future snow accumulation, rather than historic. The dataset used in these experiments was limited to Greenland, and only measured twenty ice layers. It is likely possible to generalize this model onto other polar regions, such as Antarctica, or use data with a much larger depth and thus number of ice layers. The inclusion of physical ice properties, more advanced machine learning techniques, and a deeper hyperparameter search may also serve to produce even better results.

\section{Acknowledgements}
This work is supported by NSF BIGDATA awards (IIS-1838230, IIS-1838024), IBM, and Amazon. We acknowledge data and data products from CReSIS generated with support from the University of Kansas and NASA Operation IceBridge.

%\vfill
%\pagebreak

\vfill
\pagebreak
\newpage

\bibliographystyle{IEEEbib}
\bibliography{references}

\end{document}